\documentclass[aps,prx,reprint,floatfix,superscriptaddress]{revtex4-2}

\usepackage{graphicx}
\usepackage{color}
\usepackage{amsmath}
\usepackage{amssymb}
\usepackage{braket}
\usepackage{ulem}
\usepackage{xcolor}
\usepackage{lineno}
\usepackage{hyperref}
\begin{document}

\title{Mean-Field Dynamics of Chain-of-Thought Reasoning in Large Language Models}

\author{Hao \surname{Ai}}
\altaffiliation[aihao.phys@gmail.com]{}
\affiliation{Tsinghua University, Beijing 10084, China.}

\date{\today}

\begin{abstract}
Large language models (LLMs) with chain-of-thought reasoning have been widely applied in recent years, and theoretical explanations of their behavior may help deepen our understanding and guide model optimization. In this study, we introduce a framework that seeks statistical regularities and theoretical interpretations in LLM reasoning without simplifying the model architecture or making analogies to existing physical systems. We formulate LLM reasoning as a guided discovery process on a clue graph, and derive a one-dimensional ordinary differential equation for the fraction of discovered clues using the mean-field approximation. Experimentally, clue tokens are identified using the normalized surprisal of a student LLM on the outputs of a teacher LLM, and statistical regularities are obtained by averaging over many reasoning chains of thought. Our experiments show that the resulting statistical regularities are reproducible within the same dataset and can be fitted by the solving the proposed theoretical equation.
\end{abstract}

\maketitle

\textit{Introduction--} Chain-of-thought-based large language models (LLMs) have recently demonstrated strong reasoning capabilities, with performance approaching the human level on selected tasks such as mathematical reasoning~\cite{wei2022chain,lewkowycz2022solving,wang2022self,guo2025deepseek}, competitive programming~\cite{guo2025deepseek,el2025competitive}, medical question answering~\cite{singhal2023large,singhal2025toward}, and professional or academic examinations~\cite{chung2024scaling,wang2024mmlu,shetty2025advanced}. Understanding the internal mechanisms or statistical regularities underlying LLMs~\cite{meng2022locating,wang2022interpretability,kaplan2020scaling,schaeffer2023emergent,belkin2019reconciling,vzunkovivc2024grokking} may provide useful guidance for model optimization, efficiency improvement, and cost reduction. It has therefore become an important direction in theoretical research.

Existing theoretical studies have explored deep learning systems such as LLMs from physics-inspired perspectives. One common approach is to start from first principles. Unlike most physical systems, the internal architecture of a deep learning model is fully known, allowing researchers to formulate interpretable theories based on the elementary architecture of neural networks~\cite{bahri2020statistical,roberts2022principles,NEURIPS2024_20fdaf67,olsson2022context,von2023transformers}. Nevertheless, the architectures of modern LLMs are already highly complex, making it difficult for such theories to account for all of their specialized structures. In practice, omitting architectural components such as residual connections~\cite{he2016deep}, multi-head attention~\cite{voita2019analyzing}, or MoE~\cite{fedus2022switch} can substantially reduce LLM performance, thereby limiting the universality of these theories based on simplified architectures. Another approach is to compare LLMs with well-studied physical systems, such as spin glasses~\cite{li2025spin}, systems near phase transitions~\cite{aoyama2025language,cui2024phase}, and BKT-type statistical-mechanical models~\cite{PRE_BKT}. However, these physical systems were not built to describe LLMs, and the physical concepts imported into LLM studies are often difficult to define rigorously. This weakens the rigor of such theoretical approaches.

Rather than decomposing LLMs from first principles, this work treats them at the level of collective behavior and seek statistical regularities in their reasoning processes. Instead of comparing LLMs with existing physical systems, we seek to construct a theoretical model tailored to LLMs. Using a mean-field approximation, this theory yields an ordinary differential equation that captures the statistical patterns observed in experiments.

Theoretically, we formulate chain-of-thought reasoning in LLMs as a clue discovery process. We assume that solving a problem requires a set of clues, each of which is either known or unknown. During reasoning, the LLM progressively turns unknown clues into known ones. The clues form a directed acyclic clue graph, in which known upstream clues facilitate the discovery of downstream clues. Within a mean-field approximation, we derive an ordinary differential equation for the time evolution of the fraction of known clues. We refer to this equation as the guided discovery equation. Its solution gives the time-dependent clue discovery rate.

We validate the theory experimentally by using a stronger teacher LLM to generate chains of thought and a student LLM to scan them. The observable for clues is the token-level normalized surprisal, which measures the capability gap between the two LLMs and therefore implicitly identifies key clues for solving the problem. By collecting a large number of chains of thought and averaging the resulting clue discovery rate curves, we obtain statistical patterns that demonstrate two central results: 1) under the same dataset and the same LLMs, the clue discovery rate follows reproducible statistical regularities; and 2) the averaged clue discovery rate can be partially fitted by the guided discovery equation. These findings indicate that LLM reasoning can be treated as a physical system with statistical regularities, and that such regularities can, in certain regimes, be captured by a simple equation.

\begin{figure}[tbp]
	\centering
	\includegraphics[width=1\linewidth]{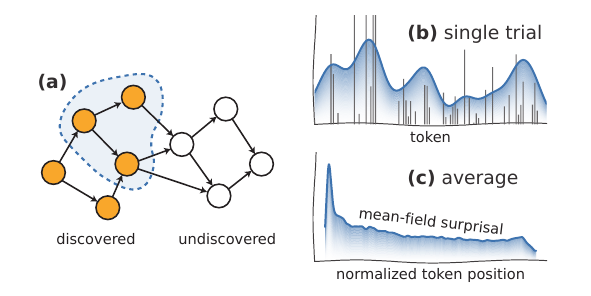}\\
	\caption{
		(a) Schematic of clue discovery. A set of clues, shown as circles, forms a directed acyclic graph. Orange-filled and empty circles denote known and unknown clues, respectively. Known clues can guide the discovery of unknown ones, but only those within the attention window are attended to. The attention window is marked by the blue dashed circle. (b) Discovery rate for a single chain of thought. Tokens with normalized surprisal above a threshold are identified as clue tokens. The vertical lines show their normalized surprisal, and the blue solid line shows the resulting discovery rate curve. (c) Average discovery rate curve over many chains of thought, revealing a statistical regularity.
	}\label{fig1}
\end{figure}

\textit{Chain-of-thought reasoning as clue discovery--} We formulate chain-of-thought reasoning in LLMs as guided clue discovery on a clue graph, as shown in Fig.~\ref{fig1}(a). Solving a problem is assumed to requires to know $N$ clues, whose dependencies form a directed acyclic graph. The LLM acts as a discovery agent on this graph, progressively converting unknown clues into known ones. The time variable of this process corresponds to the token position in the reasoning chain. Each clue has a binary state $X_i\in\{0,1\}$, where $0$ denotes unknown and $1$ denotes known. For each unknown clue $i$, its state may switch from $0$ to $1$ within a small interval $dt$ with a certain probability. The discovery probability depends on how many of its nearest upstream neighbors are already known. Since the upstream neighbors are the clues that can directly support the inference of clue $i$, the guided discovery rate of clue $i$ increases with the number of such upstream clues that have already been discovered.

%===================================================================
%fig1
%===================================================================

In addition, although an LLM can in principle access the full context when generating each new token, its attention mechanism aggregates contextual information through weighted averages over attention scores~\cite{vaswani2017attention}. As a result, the LLM may not be able to fully, uniformly, and comprehensively use all known clues. We therefore introduce an attention window into our theory. At each time, the agent attends only to a finite subset of the known clues, and only those attended clues can contribute to the guided discovery of new clues. For a typical unknown clue $i$, we assume that each known nearest upstream clue enters the attention window with probability $\rho$. The introduction of $\rho$ implicitly assumes that the LLM has the ability to select relevant clues from a large number of known clues. In contrast, if the LLM agent can only randomly sample from the known clues, the selection process reduces to a hypergeometric distribution. Specifically, suppose that there are $M$ known clues in total, among which $c$ are nearest upstream neighbors of clue $i$. If the agent randomly draws $k$ clues from the $M$ known clues into the attention window, then the guided discovery rate of clue $i$ depends on the number $r$ of selected clues that are also nearest upstream neighbors of clue $i$.

\textit{Mean-field theory--} For an individual problem, the clue discovery graph may contain substantial randomness, and the discovery trajectory of a single chain of thought may also be highly uncertain. Nevertheless, when a large number of samples are collected and averaged, the resulting behavior can exhibit certain regularities. We therefore introduce a mean-field approximation, a standard approach in statistical physics for reducing many-body interactions to an effective averaged field~\cite{chaikin1995principles}. First, we assume that every clue has the same number of nearest upstream clues, denoted by $d$. Second, if $M$ of the $N$ clues are known at a certain time, then for a representative unknown clue $i$, each of its nearest upstream clues is known with probability $M/N$. Third, as discussed earlier, a known nearest upstream clue of $i$ is attended to with probability $\rho$. As a result, under the mean-field approximation, the guided discovery process can be formulated through a binomial distribution, i.e., the guided discovery rate of clue $i$ is determined by the number $r$ of its $d$ nearest upstream clues that are both known and selected into the attention window, and each upstream clue satisfies this condition with probability $\rho m$.

In addition to this guided discovery effect, we further account for the accidental discovery of clues, which represents the LLM’s prior or intrinsic knowledge of the corresponding dataset. Consider that at most one new clue can be discovered at any moment, the clue discovery process can be reduced to an one-dimensional ordinary differential equation,
\begin{equation}\begin{aligned}\label{eq:clue-discovery}
		\frac{dm}{dt} & = (1-m)\big[\epsilon+\beta G(m)\big], \\
		G(m) & = \sum_{r=0}^{d} f(r)\binom{d}{r}(\rho m)^r(1-\rho m)^{d-r},
\end{aligned}\end{equation}
where $m$ is the dependent variable and represents the proportion of known clues, namely $m = M/N$. $\epsilon$ denotes the coefficient of accidental discovery, while $\beta$ denotes the coefficient of guided discovery. The function $f(r)$ is the guided discovery kernel. It describes the dependence of the guided discovery probability of an unknown clue $i$ on the number $r$ of upstream clues that are both known and included in the attention window. It is typically chosen as a monotonically increasing function. In the subsequent experiments, we take $f(r) = (r/d)^{0.1}$, which grows rapidly when $r$ is small and then gradually approaches saturation.

\textit{Observable: normalized surprisal--} We next introduce the experimental setup for validating the theory, together with the main observable, normalized surprisal. In the experiment, a strong teacher LLM is asked to repeatedly answer multiple questions from a dataset, thereby producing a large collection of chains of thought. We then use a weaker student LLM to scan the chains of thought generated by the teacher LLM, and identify the tokens that are difficult for the student LLM to predict. Since the teacher LLM generally performs much better than the student LLM, such hard-to-predict tokens in the teacher LLM’s reasoning process can be viewed as the crucial elements that guide the LLM toward the better answers. Accordingly, they correspond to the clues in our theory, and we call them clue tokens. Introducing both a teacher LLM and a student LLM is necessary for obtaining surprisals in our setting, and the framework is also broadly used in tasks such as knowledge distillation~\cite{hsieh2023distilling,kim2016sequence} and weak-to-strong generalization~\cite{burns2023weak}.

The surprisal $s_t$ reflects the degree to which the student LLM is surprised by the $t$-th token in the teacher LLM’s chain of thought~\cite{shannon1948mathematical}. It is defined as
\begin{equation}
	s_t	= -\log p_{\theta}\left(x_t \mid \mathcal{C}, x_{<t}\right),
\end{equation}
where $x_t$ is the $t$-th token in the teacher LLM's chain of thought, $x_{<t}$ denotes all previous tokens before position $t$, $\mathcal{C}$ denotes the prompt and problem context, and $p_{\theta}$ is the next-token probability distribution predicted by the student LLM. However, surprisal itself does not fully reflect the capability gap between the two LLMs. If a token receives a high surprisal under the student LLM, this does not necessarily mean that the student model is incapable of generating that token. It may simply be that, due to the sentence structure, the token at that position is inherently uncertain. For example, there are often many possible choices for the first token of a sentence, so its surprisal is naturally high. To address this issue, we perform z-score normalization on surprisal. Using the student LLM’s forward predictive entropy and predictive varentropy~\cite{li2025entropy,ahmed2026logitscope}, we obtain the normalized surprisal $z_t$,
\begin{equation}
	z_t = \frac{s_t - H_t}{\sqrt{V_t}},
\end{equation}
where
\begin{equation}
	H_t	= -\sum_{v\in\mathcal{V}} p_{\theta}\left(v \mid \mathcal{C}, x_{<t}\right) \log p_{\theta}\left(v \mid \mathcal{C}, x_{<t}\right),
\end{equation}
and
\begin{equation}
	V_t	= \sum_{v\in\mathcal{V}} p_{\theta}\left(v \mid \mathcal{C}, x_{<t}\right) \left[ -\log p_{\theta}\left(v \mid \mathcal{C}, x_{<t}\right) - H_t \right]^2.
\end{equation}
Here, $H_t$ is the forward prediction entropy of the student model, which is also the expectation of surprisal. It reflects the semantic uncertainty of the student model during prediction. The predictive varentropy $V_t$ is the variance of surprisal, and reflects how concentrated the predictive probability distribution is.

Finally, we obtain the statistical regularities of clues from normalized surprisal. For each individual chain of thought produced in a specific answer, we record the normalized surprisal at every token position. A token with a larger normalized surprisal is more likely to serve as a clue token. For practical simplicity, we introduce a fixed threshold $\lambda$ to determine whether each token is a clue token. Although this thresholding procedure can be rough for a single chain of thought, it is reasonable at the statistical level. In this way, normalized surprisal is transformed into a binary variable, $\hat{z}_t=\Theta(z_t-\lambda)$, where $\hat{z}_t$ equals $1$ if the token is a clue token and equals $0$ otherwise. In the above guided discovery theory, the core observable is the clue discovery rate, defined as the number of clues discovered within a short time interval. In the experiment, this quantity corresponds to the number of clue tokens in a local neighborhood around a given token. For smoothness, we use a Gaussian kernel to count the number of clues within such a local neighborhood,
\begin{equation}
	\tilde{z}_t	= \frac{ \sum_{\tau=1}^{T} K_{\sigma}(t-\tau)\hat{z}_{\tau} }{ \sum_{\tau=1}^{T} K_{\sigma}(t-\tau)	},
	\quad
	K_{\sigma}(t-\tau) = \exp\left[-\frac{(t-\tau)^2}{2\sigma^2}\right],
\end{equation}
where $T$ is the length of the chain of thought, and $\sigma$ is the bandwidth of the Gaussian smoothing kernel. At this point, we have obtained the clue discovery rate curve for a single chain of thought, as shown in Fig.~\ref{fig1}(b). The clue discovery rate curve of a single chain of thought appears random, but the average of a large number of such curves exhibits statistical regularities, as shown in Fig.~\ref{fig1}(c). Specifically, we normalize the horizontal coordinate of each clue discovery rate curve to the interval from $0$ to $1$ according to the token position, and then average the normalized curves across many chains of thought. The resulting averaged curve gives the clue discovery rate in the statistical sense, and corresponds to the clue discovery rate $dm/dt$ obtained by solving the clue discovery equation Eq.~(\ref{eq:clue-discovery}) under the mean-field approximation.

\textit{Experimental results--} In our experiments, Qwen3-Max serves as the teacher LLM and Qwen3-8B serves as the student LLM~\cite{yang2025qwen3}. Experiments are performed on four textual reasoning datasets, namely MuSR~\cite{sprague2024musr}, CLUTRR~\cite{sinha2019clutrr}, StrategyQA~\cite{geva2021did}, and FOLIO~\cite{han2024folio}. For each dataset, we choose $100$ questions and sample $10$ independent chains of thought for each question, producing a total of $1\ 000$ chains of thought per dataset. The experiments have two main objectives. The first is to verify that, with normalized surprisal as the observable, statistical regularities can be observed. The second is to adjust the model parameters and show that the mean-field clue discovery equation Eq.~(\ref{eq:clue-discovery}) is able to fit the empirical statistical curves.

%===================================================================
%fig2
\begin{figure}[tbp]
	\centering
	\includegraphics[width=1\linewidth]{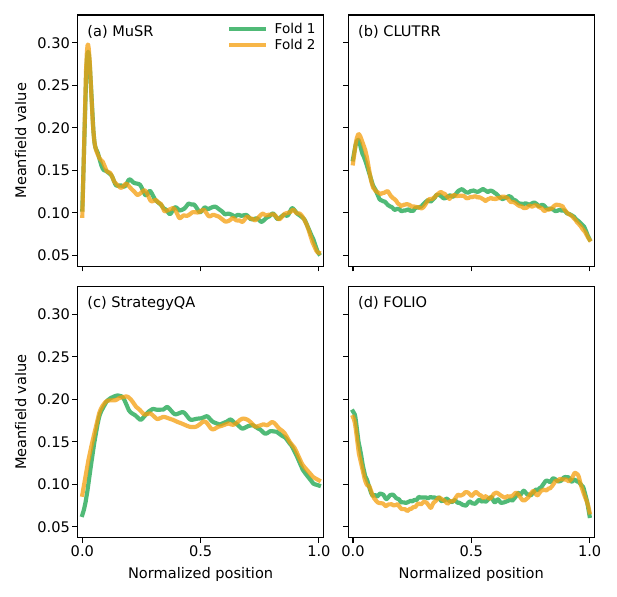}\\
	\caption{
		Two-fold experiments demonstrate the existence of the statistical regularities. The four subplots correspond to the four datasets, namely MuSR, CLUTRR, StrategyQA, and FOLIO. For each dataset, the generated chains of thought are divided by question into two non-overlapping subsets. The clue discovery rates are then computed from normalized surprisal for the two folds, shown in orange and green. The results from the two folds agree well with each other.
	}\label{fig2}
\end{figure}
%===================================================================

%===================================================================
%fig2-glm
\begin{figure}[tbp]
	\centering
	\includegraphics[width=1\linewidth]{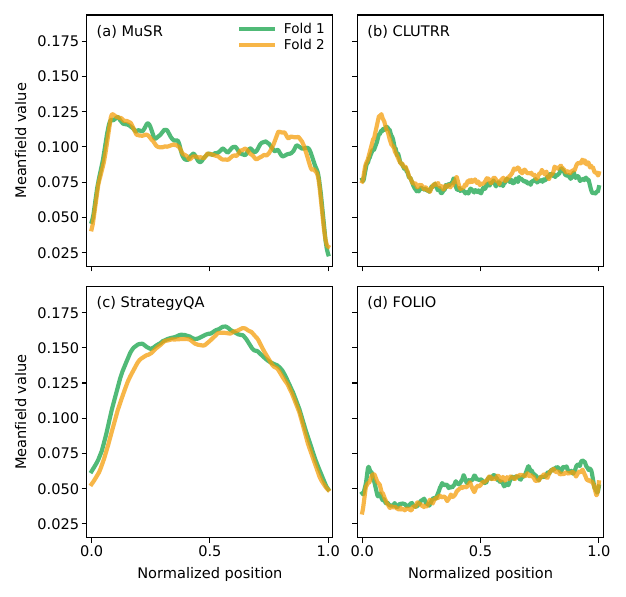}\\
	\caption{
		Two-fold experiments using GLM-4.7 as the teacher. The student LLM is kept to be Qwen3-8B. The subplots indicate that the statistical regularities hold for different teacher LLMs, but the resulting discovery rate curves are different between Qwen3-Max and GLM-4.7.
	}\label{fig2_glm}
\end{figure}
%===================================================================

To demonstrate that the surprisal-based clue discovery rate $\tilde{z}_t$ has statistical regularities, we divide the $100$ problems in each dataset into two non-overlapping parts, each containing $50$ problems, denoted as fold-1 and fold-2. We then compute the averaged $\tilde{z}_t$ curves for the two folds separately, as shown in Fig.~\ref{fig2}. As a result, the $\tilde{z}_t$ curves from the two folds exhibit strong consistency within each dataset. This indicates that the clue discovery rate $\tilde{z}_t$ indeed contains certain statistical regularities, rather than being a randomly varying curve.

%===================================================================
%fig3
\begin{figure}[tbp]
	\centering
	\includegraphics[width=1\linewidth]{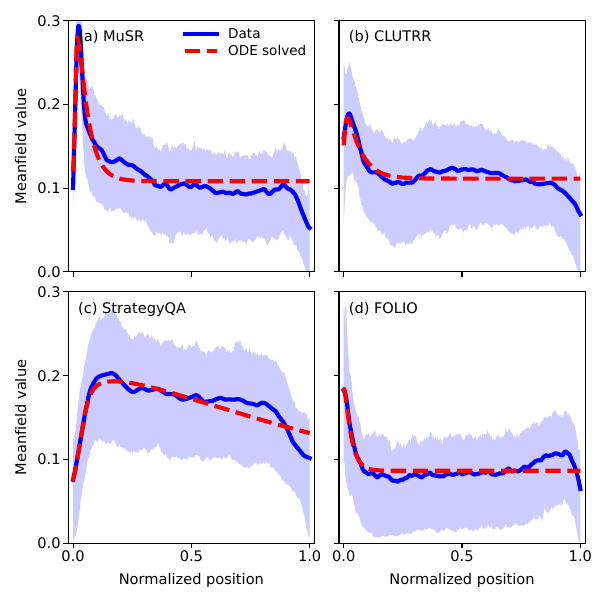}\\
	\caption{
	Fitting the theory to the experiments. The four subplots correspond to the four datasets. In each subplot, the blue solid line denotes the experimentally measured average clue discovery rate, and the blue shaded region spans the first to the third quartile. The red dashed line shows the theoretical fit. The theory agrees well with the experimental results in the first half of the reasoning process.
	}\label{fig3}
\end{figure}
%===================================================================

%===================================================================
%fig3-glm
\begin{figure}[tbp]
	\centering
	\includegraphics[width=1\linewidth]{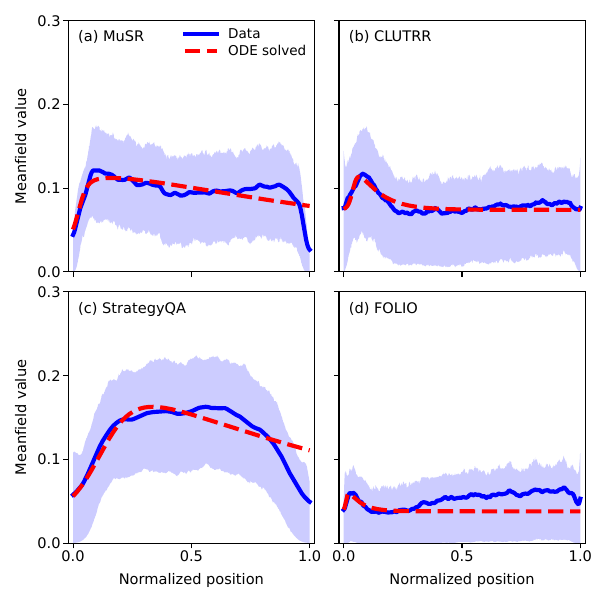}\\
	\caption{
		Fitting the theory to the experiments. GLM-4.7 serves as the teacher LLM, and Qwen3-8B is the student LLM. The theory also agrees well with the experimental results in the first half of the curve.
	}\label{fig3_glm}
\end{figure}
%===================================================================

Next, to validate the theoretical modeling, we tune the hyperparameters of the guided discovery equation so that the solved $dm/dt$ fits the experimental clue discovery rate curve of each dataset. Since a small number of clue tokens may be provided by the question or the prompt, the initial value $m(0)$ of the equation is also treated as a tunable hyperparameter, and it satisfies $0<m(0)\ll 1$. In addition, we apply a linear transformation to the $dm/dt$ obtained by solving the equation in order to fit the experimental result, namely
$ \tilde{z}_t \sim a[dm(t)/dt]+b, $
where $m(t)$ is the solution of the equation. This is because the theoretical equation gives the discovery rate of the fraction $m$ of known clues, whereas the experimental observable is the discovery rate of the number of clue tokens. Therefore, we multiply $dm/dt$ by a coefficient $a$ to adjust the scale. Moreover, because the student LLM is still less capable than the teacher LLM in extracting existing clues from the context, even already discovered clues may still produce high surprisal under the student model with some probability. We therefore introduce a bias term $b$ to reduce the influence of this mechanism. The fitting results on the four datasets are shown in Fig.~\ref{fig3}. It can be seen that the theory proposed in this study can fit the experimental results well in the first half of the evolution, which demonstrates the validity of the theory within a certain range.

In addition, we also conduct experiments using GLM-4.7~\cite{zeng2025glm} as the teacher LLM, and the corresponding results are reported in the Fig.~\ref{fig2_glm} and Fig.~\ref{fig3_glm}. 
% In addition, we resample the chains of thought according to their lengths and then take a weighted average, which further enhances the cross-dataset consistency of the results. The details are also included in the Supplementary Material~\cite{supply}.

\textit{Conclusion and limitations--} In summary, this study focuses on a theoretical interpretation of LLM chain-of-thought reasoning. Our theory neither breaks down the key components of LLMs to obtain an interpretable reduced model, nor explains LLMs by comparison with well-studied physical systems. Instead, this study offers a new perspective. It seeks to identify statistical regularities through experimental design, metric transformation, and averaging over a large number of samples. We then regard these statistical regularities as the solution of an underlying differential equation, construct a theoretical model, derive the differential equation through a mean-field approximation, and tune its hyperparameters so that the theory agrees with the experiments.

Specifically, this study regards chain-of-thought reasoning in LLMs as the guided discovery of clues. We introduce the clue graph and the attention window, and derive the clue discovery equation based on a mean-field approximation. In the experiments, we identify key clues through the capability gap between teacher and student LLMs, use normalized surprisal to characterize clue tokens, and then obtain the discovery rate curve statistically. The experimental results demonstrate that the curves are consistent across samples from the same dataset, confirming the existence of statistical regularities. With appropriate hyperparameters, the theoretical equation can fit the experimental results within a certain regime, indicating that the proposed theory at least partially captures the physical regularities underlying the real system.

As an early attempt, this study still has several limitations. First, both the theoretical modeling and the experimental setup involve many hyperparameters. Although each hyperparameter is chosen in a reasonable way, this reduces the simplicity and generality of the theory. Second, the experimental results indicate that the statistical regularities of the clue discovery rate do not exhibit consistency across different datasets and models. This may arise from differences in reasoning paths across datasets and differences in the internal knowledge of the LLMs. Finally, the experiments depend on two LLMs, a teacher and a student, which may introduce further uncontrolled factors affecting the universality of the regularities. In future research, we will look for experimental settings and observables that rely only on a single LLM, while being more interpretable and more universal, and we will develop theories to explain them.

\bibliography{ref}

\end{document}